%% file: main.tex
\documentclass[runningheads]{llncs}

\usepackage{eccv}

\usepackage{eccvabbrv}

\usepackage{graphicx}
\usepackage{booktabs}

\usepackage[accsupp]{axessibility}  

\usepackage{hyperref}

\usepackage{orcidlink}

\begin{document}

\title{Grounded Forcing: Bridging Time-Independent Semantics and Proximal Dynamics in Autoregressive Video Synthesis} 

\titlerunning{Grounded Forcing}

\author{Jintao Chen\inst{1,2*} \and
Chengyu Bai\inst{1,2*} \and
Junjun Hu\inst{2}$^{\ddagger}$ \and
Xinda Xue\inst{1,2} \and
Mu Xu\inst{2} 
}


\institute{Peking University, \and
Alibaba \\
\email{\{cjt, chengyu.bai\}@stu.pku.edu.cn}
}

\maketitle

\begingroup
\renewcommand{\thefootnote}{}
\footnotetext{\textsuperscript{*} Equal contribution. \textsuperscript{$\ddagger$} Corresponding author.}
\addtocounter{footnote}{-1}
\endgroup

\begin{figure}[ht!]
    \centering
    \vspace{-15pt}
    \includegraphics[width=\linewidth]{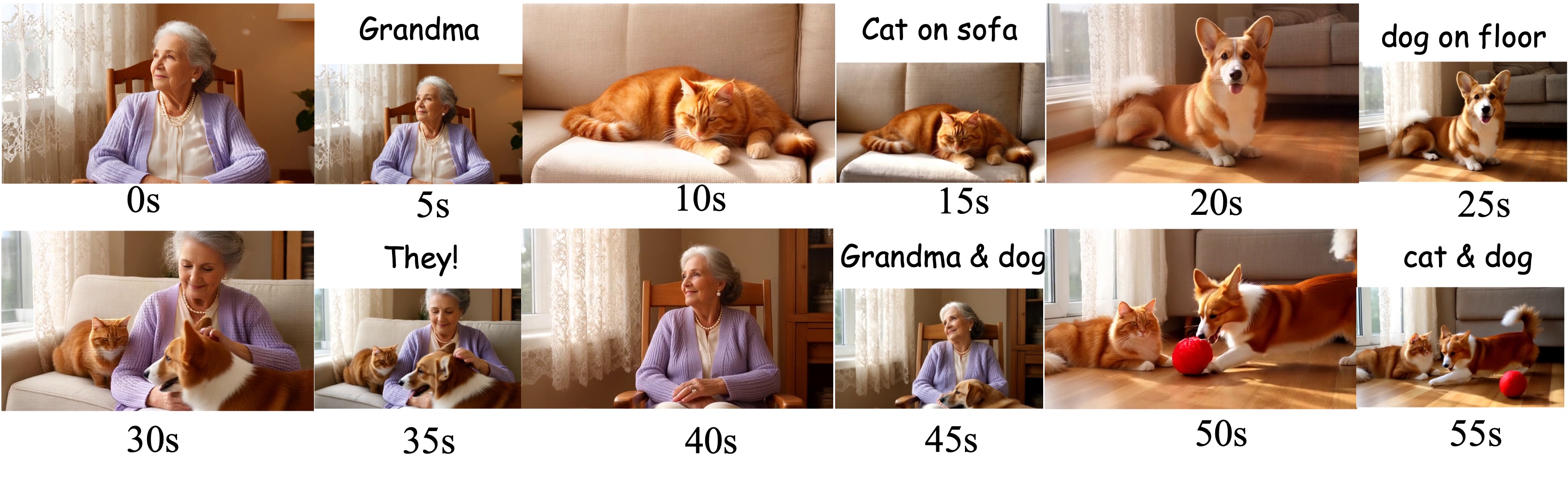}
    \vspace{-25pt}
    \caption{\textbf{Grounded Forcing for Long-Horizon Interactive Video Generation.} Our method generates a coherent 1-minute video with multiple characters (Grandma, cat, and dog) across diverse scenes, maintaining long-term identity consistency while supporting dynamic prompt switching and multi-shot transitions.}
    \label{fig:teaser}
    \vspace{-35pt}
\end{figure}

\input{sections/0_abs}
\input{sections/1_introduction}
\input{sections/2_related_work}

\input{sections/3_preliminaries}
\input{sections/3_methods}

\input{sections/4_experiments}
\input{sections/5_conclusion}
%
%
\bibliographystyle{splncs04}
\bibliography{main}

\input{sections/X_supp}

\end{document}

%% file: sections/0_abs.tex
\begin{abstract}
Autoregressive video synthesis offers a promising pathway for infinite-horizon generation but is fundamentally hindered by three intertwined challenges: semantic forgetting from context limitations, visual drift due to positional extrapolation, and controllability loss during interactive instruction switching. Current methods often tackle these issues in isolation, limiting long-term coherence. We introduce Grounded Forcing, a novel framework that bridges time-independent semantics and proximal dynamics through three interlocking mechanisms. First, to address semantic forgetting, we propose a Dual Memory KV Cache that decouples local temporal dynamics from global semantic anchors, ensuring long-term semantic coherence and identity stability. Second, to suppress visual drift, we design Dual-Reference RoPE Injection, which confines positional embeddings within the training manifold while rendering global semantics time-invariant. Third, to resolve controllability issues, we develop Asymmetric Proximity Recache, which facilitates smooth semantic inheritance during prompt transitions via proximity-weighted cache updates. These components operate synergistically to tether the generative process to stable semantic cores while accommodating flexible local dynamics. Extensive experiments demonstrate that Grounded Forcing significantly enhances long-range consistency and visual stability, establishing a robust foundation for interactive long-form video synthesis.
\vspace{-10pt}
\keywords{Autoregressive video synthesis \and long-term coherence}
\end{abstract}

%% file: sections/1_introduction.tex
\section{Introduction}
The pursuit of interactive, infinite-length video generation represents a pivotal step toward immersive world simulators. While non-autoregressive video synthesis~\cite{ltx,cogvideox,wan,hunyuanvideo,moviegen,seawead} has achieved remarkable fidelity via Diffusion Transformers~\cite{Dit}, their reliance on bidirectional attention precludes KV caching, resulting in redundant computation and prohibitive inference latency. Consequently, enabling real-time streaming necessitates a paradigm shift toward autoregressive architectures~\cite{CausVid,SelfForcing,teng2025magi,chen2025skyreels,framepack,diffusionforcing}. Such causal frameworks inherently align with temporal progression, facilitating low-latency synthesis and dynamic interaction. However, extending generation from short clips to continuous, long-horizon streams introduces fundamental instabilities that remain unresolved. Specifically, autoregressive video synthesis is hindered by three interconnected challenges: \textbf{semantic forgetting} due to context window limitations, \textbf{visual drift} caused by positional extrapolation, and \textbf{controllability degradation} during dynamic instruction switching.

\begin{figure}[t]
    \centering
    \includegraphics[width=\linewidth]{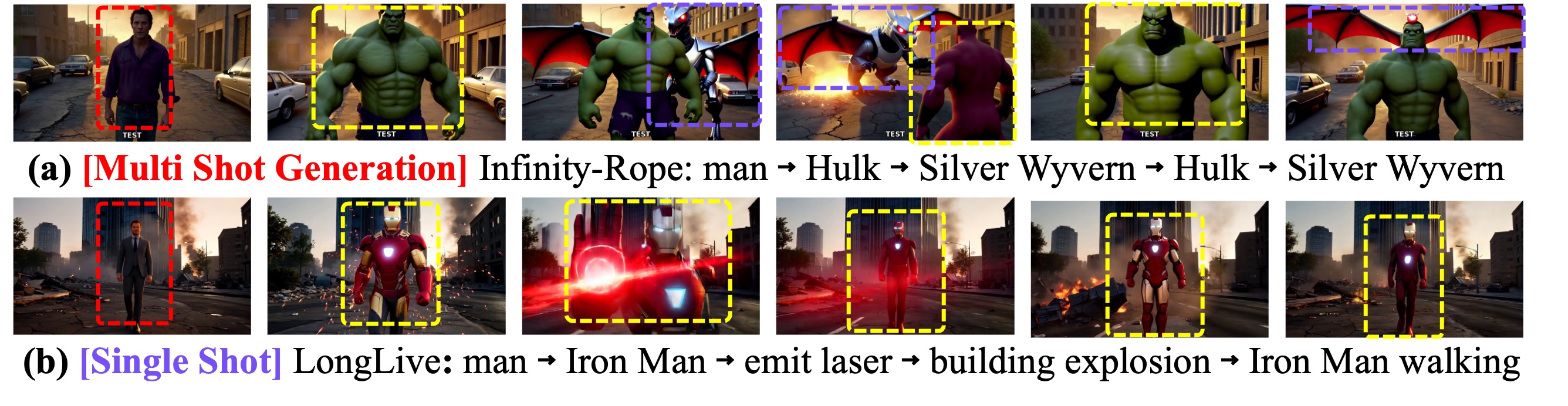}
    \vspace{-15pt}
    \caption{\textbf{Limitations of Existing Methods.} 
    (a) \textbf{Infinity-RoPE in Multi-Shot Generation:} The model remains fixated on the initial frame (man, red box) and fails to preserve newly introduced entities (Hulk in yellow, Silver Wyvern in purple), demonstrating \textbf{semantic forgetting}.
    (b) \textbf{LongLive in Continuous Generation:} Rigid anchoring to initial semantics causes visual inconsistencies during semantic transitions, hindering \textbf{semantic evolution} and controllability.}
    \label{fig:motivation}
    \vspace{-15pt}
\end{figure}

While recent advancements~\cite{framepack,SelfForcing,SelfForcing++,rollingforcing,InfinityRope,longlive,contextforcing,lu2025reward,guo2025end} have sought to mitigate instabilities in autoregressive video diffusion, these approaches typically address the aforementioned challenges in isolation. To combat \textit{semantic forgetting}, prevailing methods~\cite{longlive, InfinityRope,rollingforcing} anchor attention to the first frame to preserve initial semantics throughout the generation process. However, this rigid reliance on the initial frame is inherently limited in dynamic scenarios. As illustrated in Figure~\ref{fig:motivation}, the model remains fixated on the first frame (e.g., man) rather than adapting to emerging semantics (e.g., Hulk, Silver Wyvern), thereby failing to \textbf{incorporate new semantic content}. Consequently, the semantics of newly introduced content remain forgotten. Moreover, this static anchor unduly constrains semantically independent content, leading to abrupt visual jumps or semantic contamination from the initial frame, thus \textbf{impeding semantic evolution}. To mitigate controllability degradation, existing methods~\cite{longlive, InfinityRope} propose refreshing the KV Cache with a new prompt to enable prompt switching. However, they strictly require independence between prompts and struggle to maintain consistency in scenarios with drastic contextual shifts.

To bridge this gap, we present \textbf{Grounded Forcing}, a comprehensive framework designed to anchor time-independent semantics while accommodating proximal dynamics. The name reflects our core philosophy: \textit{Forcing} denotes the autoregressive generation process, while \textit{Grounded} signifies our mechanism to tether the generative trajectory to stable semantic anchors, preventing it from drifting into incoherence. Unlike purely inference-time adaptations, our approach introduces structural innovations that operate synergistically to address the forgetting-drifting-controllability triad.

First, to combat semantic forgetting, we propose a \textbf{Dual Memory KV Cache} that architecturally decouples memory storage into Local Temporal Memory (LTM) for high-frequency motion and Global Consistency Memory (GCM) for persistent semantic concepts. This ensures that global semantic integrity and identity features remain anchored regardless of temporal depth. Second, to suppress visual drift, we design \textbf{Dual-Reference RoPE Injection}, which confines positional embeddings~\cite{su2024roformer} within the model's training manifold by assigning time-invariant indices to global anchors and relative indices to local frames. This prevents the positional degradation that typically disrupts long-term semantic coherence. Third, to resolve controllability loss, we develop \textbf{Asymmetric Proximity Recache (APR)}, a fine-grained update strategy that facilitates smooth semantic inheritance during prompt transitions. By weighting cache updates based on temporal proximity, APR allows immediate adherence to new instructions while preserving long-term context, eliminating the ``semantic shock'' of uniform cache refreshing.

These components form an interlocking system where the Dual Memory provides the structural foundation, Dual-Reference RoPE ensures positional stability, and APR enables flexible interaction. By combining these strategies, Grounded Forcing transforms standard autoregressive models into robust engines for interactive simulation. Our system operates efficiently, delivering long-term semantic coherence and identity stability suitable for real-time applications. The primary contributions of this work are:
\begin{itemize}
\item We identify the entanglement of static semantics and dynamic motion as a key bottleneck in long-form autoregressive video generation and propose \textbf{Grounded Forcing} to systematically decouple these factors.
\item We design a Dual Memory KV Cache that anchors long-term semantic coherence via Global Consistency Memory while capturing motion through Local Temporal Memory, fundamentally mitigating semantic degradation.
\item We introduce Dual-Reference RoPE Injection to prevent positional drift and Asymmetric Proximity Recache for seamless prompt switching, ensuring both visual stability and interactive controllability.
\item We demonstrate through extensive experiments that our framework achieves state-of-the-art semantic coherence and identity stability in long-form video generation while maintaining real-time performance.
\end{itemize}

%% file: sections/2_related_work.tex
\section{Related Work}

\paragraph{\textbf{Bidirectional Video Diffusion Models.}}
High-fidelity video generation is predominantly defined by bidirectional denoising diffusion models~\cite{Dit,ho2020denoising,lipman2022flow}, including Wan~\cite{wan}, HunyuanVideo~\cite{hunyuanvideo}, and Sora. These models leverage both past and future frames during denoising to achieve exceptional fidelity and have spawned numerous downstream tasks, including controllable generation~\cite{mao2026omni} and personalized generation\cite{chen2026conceptweaver,bai2025uniedit}. However, this reliance on full temporal context is incompatible with real-time streaming, where future frames are unavailable at inference. Such non-causal dependency precludes key-value (KV) caching and sequential inference, creating a fundamental barrier for low-latency applications. While training-free extensions~\cite{FreeLong,RIFLEx} offer limited horizon expansion, they often suffer from pronounced consistency degradation. 

\vspace{-10pt}

\paragraph{\textbf{Autoregressive Video Generation for Long-Horizon Synthesis.}}
To enable real-time streaming and long-horizon video generation, research has shifted toward autoregressive (AR) video diffusion models, which reformulate bidirectional diffusion into a causal framework that generates frames sequentially while leveraging KV caching for efficient inference. CausVid~\cite{CausVid} first distilled bidirectional DiTs into causal AR generators using Distribution Matching Distillation (DMD~\cite{dmd,dmd2}), compressing multi-step denoising to enable streaming synthesis. Self-Forcing~\cite{SelfForcing} further bridged the train-test gap by conditioning on self-generated frames and a rolling KV cache during training, significantly reducing exposure bias. Rolling Forcing~\cite{rollingforcing} addressed error accumulation via a rolling-window joint denoising strategy, processing consecutive frames with progressive noise levels to enable mutual refinement and suppress propagation. LongLive~\cite{longlive} aligned training and inference via a train-long-test-long strategy, employing short window attention with a frame-level sink for efficient long-range consistency.
Another critical challenge involves breaking the fixed temporal horizon of 3D-RoPE, which restricts AR models to a pre-trained maximum frame count. Self-Forcing++~\cite{SelfForcing++} extended training horizons via long rollouts but incurred prohibitive computational costs while remaining constrained by absolute ROPE time location indices. Conversely, Infinity-RoPE~\cite{InfinityRope} introduced a training-free Block-Relativistic RoPE, reformulating temporal encoding as a moving reference frame to enable infinite-horizon generation. However, as a purely inference-time adaptation, it inherits the base model's semantic drift due to the lack of training-stage semantic anchoring.

\vspace{-10pt}

\paragraph{\textbf{Interactive and Controllable Long Video Generation.}}
Interactive long video generation~\cite{zhang2025pretraining} enables user steering via streaming prompts and cinematic transitions, introducing three core challenges: prompt switching responsiveness (instant alignment with new prompts), fine-grained interactive control (adhering to new prompts without semantic contamination), and multi-shot transitions (abrupt scene cuts with global coherence). For prompt switching, KV cache reconfiguration is the primary approach: LongLive~\cite{longlive} proposes KV-recache to rebuild cache at switch boundaries; Infinity-RoPE~\cite{InfinityRope} introduces KV Flush to reset cache to two anchor frames for instant responsiveness, yet sacrifices fine-grained consistency control. For cinematic transitions, Infinity-RoPE's RoPE Cut performs discontinuous jumps in temporal RoPE coordinates to enable scene cuts, but introduces transition-edge artifacts and fails to truncate error accumulation paths.

%% file: sections/3_preliminaries.tex
\section{Preliminaries and Analysis}
\label{sec:preliminaries}

Before detailing our methodology, we formally define the autoregressive video generation framework and analyze the fundamental bottlenecks that hinder long-horizon consistency: semantic forgetting, visual drift, and controllability loss.

\subsection{Autoregressive Video Synthesis}
Let a video sequence be represented as a series of latent frames $\mathbf{X} = \{x_1, x_2, \dots, x_T\}$. In autoregressive video diffusion~\cite{SelfForcing}, the generation process is modeled as a conditional probability distribution where each frame $x_t$ is predicted based on all preceding frames $\mathbf{X}_{<t}$ and a text condition $c$:
\begin{equation}
    P(\mathbf{X}) = \prod_{t=1}^{T} P(x_t | \mathbf{X}_{<t}, c).
    \label{eq:ar_prob}
\end{equation}
To enable efficient inference, Transformer-based architectures utilize a Key-Value (KV) Cache to store historical features, avoiding recomputation of past tokens. The attention mechanism at step $t$ computes:
\begin{equation}
    \text{Attention}(Q_t, K_{<t}, V_{<t}) = \text{softmax}\left(\frac{Q_t K_{<t}^\top}{\sqrt{d}}\right) V_{<t},
    \label{eq:attention}
\end{equation}
where $Q_t, K_{<t}, V_{<t}$ are the query, key, and value projections. To incorporate temporal information, Rotary Positional Embeddings (RoPE) are typically applied to the queries and keys before caching:
\begin{equation}
    \tilde{K}_t = \text{RoPE}(K_t, t), \quad \tilde{Q}_t = \text{RoPE}(Q_t, t),
    \label{eq:rope}
\end{equation}
where $t$ denotes the absolute temporal index. While effective for short sequences, this standard formulation encounters critical failures when $T \to \infty$.

\subsection{The Triad of Challenges}

\paragraph{\textbf{1. Semantic Forgetting (The Memory Bottleneck).}}
In infinite-horizon generation, memory constraints necessitate a sliding window strategy where only the most recent $W$ frames are retained in the KV cache. While this maintains constant memory usage, it induces \textit{semantic forgetting}. Mathematically, as $t \gg W$, the attention distribution becomes independent of the initial frames $x_{1 \dots k}$ that define subject identity and scene layout:
\begin{equation}
    \lim_{t \to \infty} P(x_t | \mathbf{X}_{<t}) \approx P(x_t | \mathbf{X}_{t-W:t}),
    \label{eq:forgetting}
\end{equation}
where the global context is effectively truncated. This leads to identity swaps or background inconsistencies, as the model loses access to the semantic anchors established at the beginning of the sequence.

\paragraph{\textbf{2. Visual Drift (The Positional Bottleneck).}}
Standard RoPE mechanisms bind tokens to absolute temporal indices. During training, models observe temporal indices within a fixed range $[0, T_{\text{train}}]$. However, in streaming generation, the temporal index $t$ grows indefinitely. When $t > T_{\text{train}}$, the positional embeddings enter an out-of-distribution (OOD) regime. The rotational frequencies $e^{i m \theta}$ in RoPE become misaligned with the learned attention patterns, causing the attention scores to degenerate. This \textit{visual drift} manifests as accumulating artifacts, motion jitter, and eventual quality collapse, as the model struggles to interpret positional signals it never encountered during optimization.

\paragraph{\textbf{3. Controllability Loss (The Interaction Bottleneck).}}
Interactive video generation requires adapting to dynamic prompt switches (e.g., changing from ``walk'' to ``run''). Existing Methods ~\cite{longlive,InfinityRope} update KV Cache typically employ uniform refreshing or full flushing upon prompt changes. Let $\mathbf{K}^{\text{old}}$ and $\mathbf{K}^{\text{new}}$ denote the caches under the old and new prompts. A uniform update implies:
\begin{equation}
    \mathbf{K}' = (1 - \alpha) \mathbf{K}^{\text{old}} + \alpha \mathbf{K}^{\text{new}}, \quad \forall t,
    \label{eq:controllability}
\end{equation}
where $\alpha$ is a constant with a value of 1. This \textit{symmetric} approach fails to account for temporal proximity. Recent frames require high $\alpha$ to adhere to new motions, while distant frames require low $\alpha$ to preserve identity continuity. The lack of fine-grained control leads to \textit{semantic shock} (abrupt visual changes) or \textit{instruction ignoring} (failure to adopt new actions), severely limiting interactive usability.

%% file: sections/3_methods.tex
\section{Methods}

\subsection{Dual memory mechanism}
\begin{figure}[ht!]
    \centering
    \includegraphics[width=\linewidth]{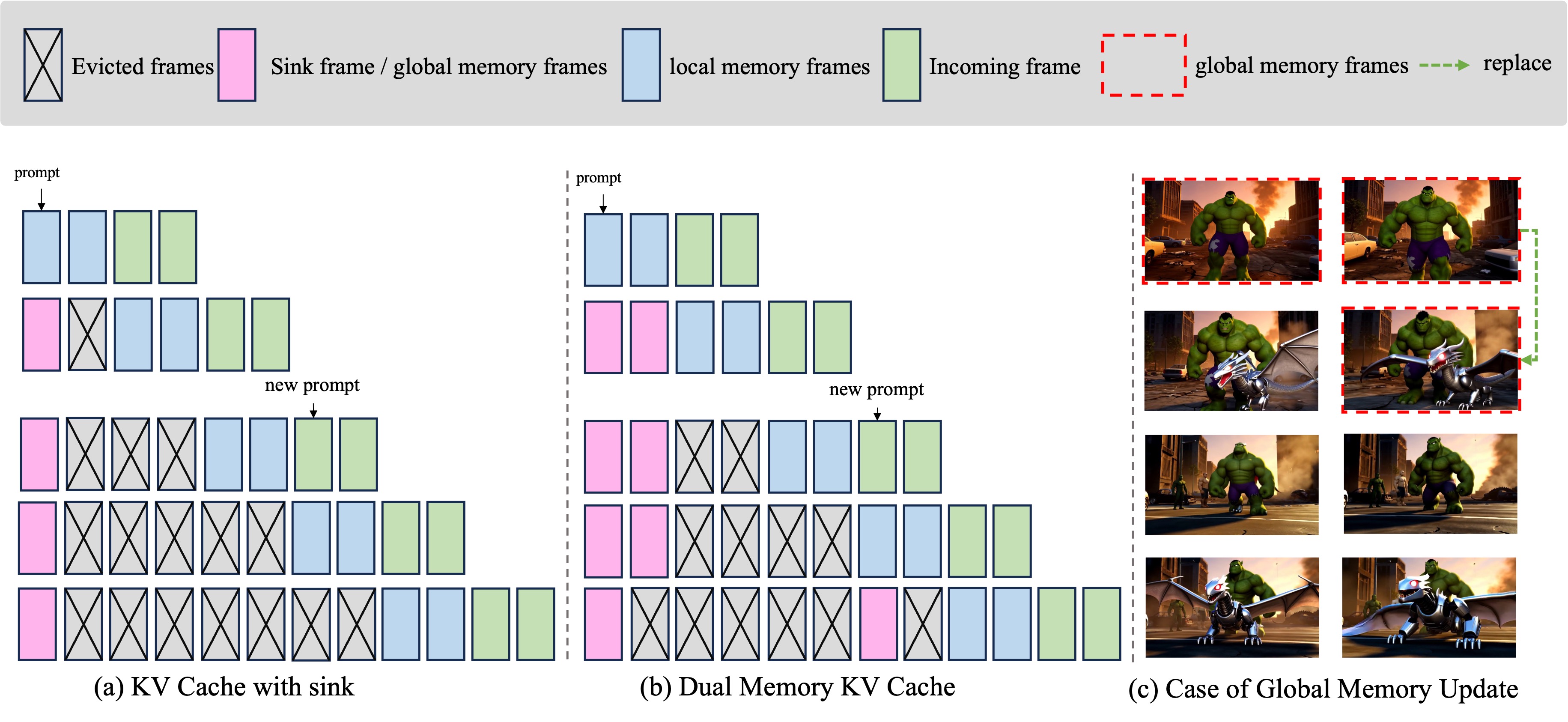}
    \caption{\textbf{Dual Memory Mechanism.} (a) Conventional KV Cache with a single sink frame, which suffers from semantic dilution over time. (b) Our Dual Memory KV Cache separates Local Temporal Memory (sliding window) from Global Consistency Memory (persistent anchors). (c) Case of Global Memory Update: Upon prompt switching, the global memory is selectively updated based on latent diversity. New frames with high semantic novelty (low similarity to GCM) replace redundant anchors (high similarity to remaining GCM), ensuring the memory spans the evolving semantic space (e.g., from Hulk to Hulk and Silver Wyvern).}
    \label{fig:dualmem}
\end{figure}

Standard autoregressive video generation typically relies on sliding window attention to manage memory constraints. While efficient, this approach inherently suffers from \textit{semantic forgetting}, as information outside the local window is permanently discarded. Recent works~\cite{longlive,InfinityRope} have attempted to mitigate this by introducing a persistent \textit{attention sink} frame. However, as illustrated in Figure~\ref{fig:dualmem}(a), a single sink frame is insufficient to capture complex, evolving semantic states over long horizons. It provides a weak global anchor that often fails to preserve subject identity or background consistency during significant scene transitions or interactive prompt switches.

To address these limitations, we propose the \textbf{Dual Memory KV Cache}, which decouples the storage of short-term dynamics from long-term semantics. As shown in Figure~\ref{fig:dualmem}(b), our cache is partitioned into two distinct components:

\begin{itemize}
\item \textbf{Local Temporal Memory (LTM):} Represented by blue blocks, this functions as a standard sliding window that stores recent frames. It captures high-frequency motion and proximal dynamics. Frames in the LTM are subject to eviction (crossed blocks) as new incoming frames (green blocks) arrive, ensuring constant memory usage for temporal processing.
\item \textbf{Global Consistency Memory (GCM):} Represented by pink blocks, this stores a sparse set of keyframes that anchor the video's semantic identity. Unlike the LTM, frames in the GCM are protected from standard sliding window eviction.
\end{itemize}

\paragraph{\textbf{Diversity-Aware Global Update.}}
The GCM is not merely a static store of initial frames; it is dynamically updated to reflect legitimate semantic shifts, particularly after prompt switches. We introduce a \textit{Diversity-Aware Update Strategy} that computes the importance of new frames and the redundancy of existing anchors based on latent similarities.

Let $z_t$ denote the latent representation of the newly generated frame at step $t$, and $\mathcal{M}_G = \{z_1, \dots, z_K\}$ be the current set of latents in the GCM. 
First, we evaluate the \textbf{importance} of $z_t$ by measuring its similarity to the existing anchors. Intuitively, a frame is important if it carries semantic information distinct from what is already stored. We define the importance score $\mathcal{I}(z_t)$ as the inverse of its maximum similarity to $\mathcal{M}_G$:
\begin{equation}
    \mathcal{S}_{max}(z_t) = \max_{z \in \mathcal{M}_G} \text{CosSim}(z_t, z), \quad \mathcal{I}(z_t) = 1 - \mathcal{S}_{max}(z_t),
\end{equation}
where $\text{CosSim}(\cdot, \cdot)$ denotes the cosine similarity. A lower similarity $\mathcal{S}_{max}$ yields a higher importance $\mathcal{I}$, indicating that $z_t$ represents a novel semantic state.

Second, to determine which anchor in $\mathcal{M}_G$ should be replaced, we assess the \textbf{redundancy} of each existing latent frame $z_i$. Crucially, to avoid self-similarity bias, we compute the similarity of $z_i$ against the \textit{remaining} set $\mathcal{M}_G \setminus \{z_i\}$ (a leave-one-out strategy):
\begin{equation}
    \mathcal{R}(z_i) = \max_{z_j \in \mathcal{M}_G, j \neq i} \text{CosSim}(z_i, z_j).
\end{equation}
A high redundancy score $\mathcal{R}(z_i)$ implies that $z_i$ is semantically close to other anchors in the GCM and is thus expendable.

Finally, we perform the update by replacing the most redundant anchor $z_{target} = \arg\max_{z \in \mathcal{M}_G} \mathcal{R}(z)$ if the new frame's importance exceeds this redundancy:
\begin{equation}
    \text{if } \mathcal{I}(z_t) > \mathcal{R}(z_{target}), \text{ then } \mathcal{M}_G \leftarrow (\mathcal{M}_G \setminus \{z_{target}\}) \cup \{z_t\}.
\end{equation}
As illustrated in Figure~\ref{fig:dualmem}(c), this mechanism ensures that the GCM evolves to cover diverse semantic states (e.g., adding the Silver Wyvern) while discarding redundant information, thereby maintaining a compact yet comprehensive semantic anchor for the video.

\subsection{Dual-Reference RoPE Injection}

A critical bottleneck in autoregressive video diffusion lies in positional encoding. Standard implementations, including Self Forcing~\cite{SelfForcing}, apply Rotary Position Embeddings (RoPE) to keys and queries \textit{before} storing them in the KV cache. While effective for short sequences, this approach locks each token to an absolute temporal index. In infinite-horizon generation, these indices grow indefinitely, then exceeding the model's trained distribution (e.g., the 81-frame limit in Wan~\cite{wan}). When the model encounters positional indices unseen during training, attention mechanisms degenerate, leading to severe visual drift and quality collapse.

\begin{figure}[ht!]
    \centering
    \includegraphics[width=\linewidth]{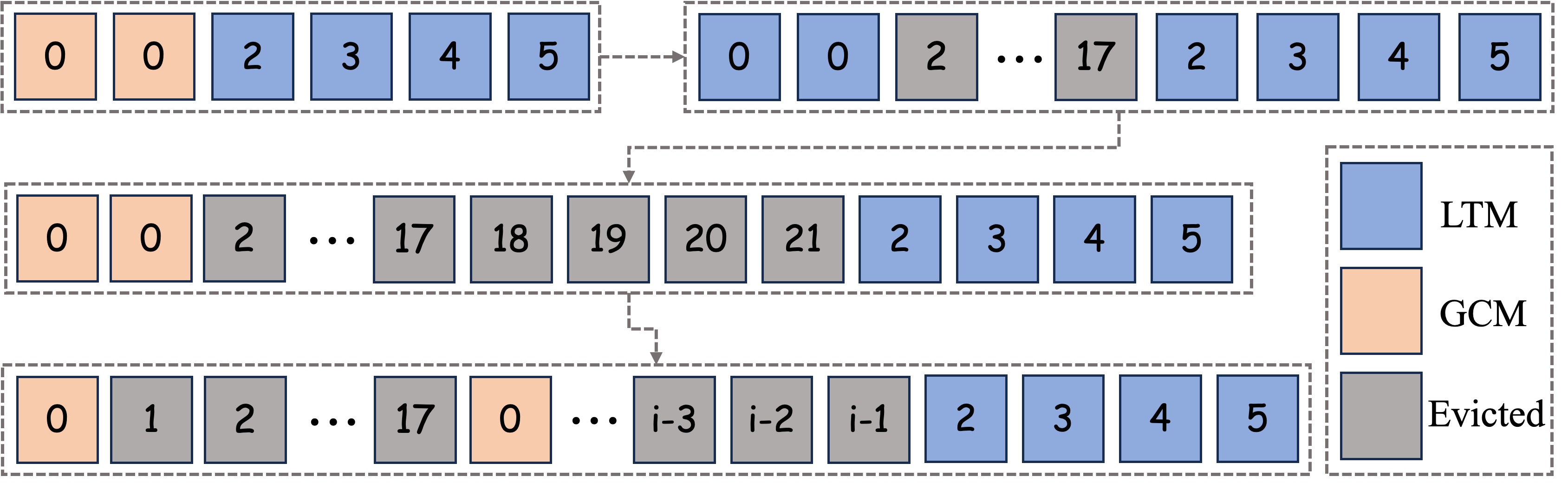}
    \caption{\textbf{Dual-Reference RoPE Injection.} Visualization of the positional encoding management over generation steps. \textbf{Orange blocks} represent Global Consistency Memory (GCM) keys, always injected with a fixed RoPE index of 0 to ensure position-agnostic semantic anchoring. \textbf{Blue blocks} denote Local Temporal Memory (LTM) keys, assigned relative indices within the model's perceptual range (e.g., $[0, 21]$). \textbf{Gray blocks} indicate evicted frames, labeled with their \textit{absolute} temporal positions (e.g., $2, \dots, i-1$) to illustrate the continuous timeline extending beyond the fixed cache window. This strategy ensures indices remain within the training distribution while preserving long-term semantics.}
    \label{fig:drrope}
\end{figure}

To overcome this, inspired by ~\cite{InfinityRope}, we introduce \textbf{Dual-Reference RoPE Injection (DR-RoPE)}. Instead of caching position-encoded keys, we store the \textit{raw} keys $\mathbf{K}_{\text{raw}}$ (pre-RoPE) in both the Global Consistency Memory (GCM) and Local Temporal Memory (LTM). Positional information is injected dynamically during the attention computation phase, allowing us to assign distinct temporal reference frames to different memory components, as illustrated in Figure~\ref{fig:drrope}.

\paragraph{\textbf{Global Consistency Memory: Time-Invariant}}
For the GCM, our goal is to maintain semantic identity regardless of the video duration. We achieve this by injecting a \textit{fixed} temporal index into the RoPE function for all keys stored in the GCM:
\begin{equation}
    \mathbf{K}_{\text{GCM}} = \text{RoPE}(\mathbf{K}_{\text{raw}}, t_{\text{global}}=0).
\end{equation}
By forcing all global memory tokens to occupy the same temporal coordinate ($t=0$), we render the GCM \textit{position-agnostic}. This ensures that the model attends to these frames as a timeless semantic reference (e.g., subject identity, core style) rather than a specific moment in time. Consequently, the global anchor remains stable and does not drift as the generation timeline extends, effectively tethering the video's semantic core.

\paragraph{\textbf{Local Temporal Memory: Relative Contextualization.}}
For the LTM, preserving motion continuity is paramount. We inject temporal indices relative to the current sliding window, ensuring they always fall within the model's optimal perceptual range. Specifically, for the Wan backbone~\cite{wan}, the model is most sensitive to temporal indices within $[0, 21]$ (latent frames). We map the local window indices to this fixed range:
\begin{equation}
    \mathbf{K}_{\text{LTM}} = \text{RoPE}(\mathbf{K}_{\text{raw}}, t_{\text{local}}), \quad t_{\text{local}} \in [0, 21].
\end{equation}
As the sliding window advances, older frames are evicted, and new frames are assigned indices within this safe bound. This prevents the positional embeddings from entering out-of-distribution regimes, ensuring that local motion dynamics remain smooth and consistent with the training distribution.

\paragraph{\textbf{Multi-Shot Generation via LTM Reset.}}
Beyond continuous generation, our framework naturally supports \textbf{Multi-Shot Generation} for cinematic storytelling involving distinct scenes. To transition between shots without visual artifacts or temporal contamination, we implement a \textit{Local Memory Reset} strategy. Upon detecting a scene transition (e.g., via a specific trigger token or a drastic prompt shift), we selectively flush the KV cache in the LTM while retaining the GCM to preserve global identity consistency.

\paragraph{\textbf{Suppressing Error Accumulation.}}
This dual-reference strategy directly mitigates error accumulation. In standard autoregressive rollout, small prediction errors compound as the temporal index grows into unseen territories, causing the model to hallucinate or lose coherence. By resetting the local temporal indices to a known safe range and anchoring global semantics to a fixed point, \textbf{Dual-Reference RoPE Injection} confines the model's operation within its high-confidence positional manifold. This design not only maintains visual continuity across long sequences but also significantly enhances generation quality by preventing the positional drift that typically plagues infinite-horizon autoregressive synthesis.

\subsection{Asymmetric Proximity Recache}

Interactive long-video generation requires the model to adapt to evolving user instructions while maintaining temporal coherence. Existing KV ReCache strategies~\cite{longlive, InfinityRope} typically employ a uniform or binary update mechanism when switching prompts: the cache is either fully flushed or refreshed equally across all timesteps. While effective for independent scene generation, this \textit{symmetric} approach fails in scenarios involving \textit{semantic inheritance}. For instance, when a prompt shifts from \textit{"A man walks"} to \textit{"Then he runs"}, the pronoun \textit{"he"} and the temporal connector \textit{"then"} rely on context from the previous prompt. A uniform cache refresh discards this linguistic dependency, leading to semantic discontinuity or abrupt visual jumps (e.g., identity changes) as the model loses track of the subject.

\begin{figure}[t]
    \centering
    \includegraphics[width=\linewidth]{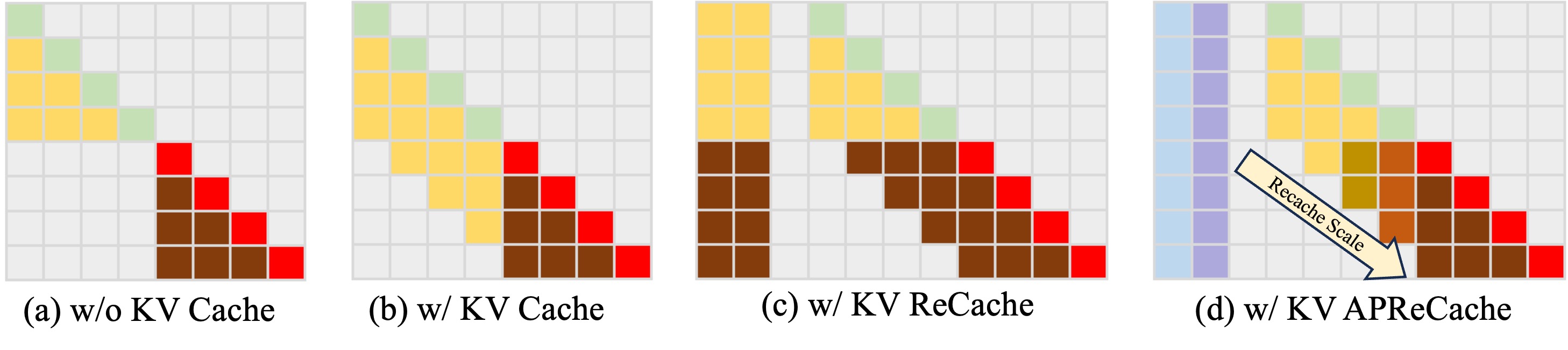}
    \caption{\textbf{Asymmetric Proximity Recache (APR).} Comparison of KV Cache strategies during prompt switching. (a-b) Standard caching fails to adapt. (c) Uniform KV ReCache refreshes all frames equally, causing semantic discontinuity. (d) Our APR applies a proximity-dependent recache scale (arrow), refreshing recent frames more aggressively to follow the new prompt while retaining distant frames to preserve semantic inheritance.}
    \label{fig:APR}
\end{figure}

To address this, we propose \textbf{Asymmetric Proximity Recache (APR)}, a fine-grained cache update strategy that balances prompt responsiveness with semantic continuity. As illustrated in Figure~\ref{fig:APR}(d), instead of applying a global refresh rate, APR computes a timestep-dependent weighting scale $\alpha_t$ for interpolating between the existing cache (carrying previous semantics) and the newly computed cache (carrying current prompt semantics).

\paragraph{\textbf{Proximity-Weighted Interpolation.}}
The core intuition of APR is that frames temporally closer to the current generation step should align more strongly with the \textit{new} prompt to ensure immediate instruction adherence, while distant frames should retain more of the \textit{old} cache to preserve long-term context. For each token $t$ in the KV cache, we compute the updated key $\mathbf{K}_t'$ and value $\mathbf{V}_t'$ as:
\begin{equation}
    \mathbf{K}_t' = (1 - \alpha_t) \mathbf{K}_t^{\text{old}} + \alpha_t \mathbf{K}_t^{\text{new}},
\end{equation}
where $\mathbf{K}^{\text{old}}$ represents the cached keys from the previous prompt phase, and $\mathbf{K}^{\text{new}}$ represents keys recomputed under the new prompt. The weighting coefficient $\alpha_t \in [0, \alpha_{\max}]$ is determined by the temporal proximity to the current generation frontier. We employ a linear decay schedule based on the temporal distance $d_t$ from the current step:
\begin{equation}
    \alpha_t = \min\left(\alpha_{\max}, 1 - \frac{d_t}{D_{\text{window}}}\right),
\end{equation}
where $D_{\text{window}}$ is the size of the recache window and $\alpha_{\max}$ limits the injection of new semantics, thereby ensuring a baseline retention of old semantics.

\paragraph{\textbf{Semantic Bridging.}}
As shown in Figure~\ref{fig:APR}, this asymmetric design creates a \textit{semantic bridge} between prompts. 
\begin{itemize}
    \item \textbf{High Proximity (Recent Frames):} A higher $\alpha_t$ allows the immediate context to rapidly adapt to the new instruction (e.g., changing motion or event), ensuring the next generated frame follows the new directive.
    \item \textbf{Low Proximity (Distant Frames):} A lower $\alpha_t$ preserves the original semantic anchors (e.g., subject identity, background) from the previous prompt. 
\end{itemize}
This gradient update prevents the \textit{semantic shock} observed in uniform ReCache methods, where the sudden loss of all prior context causes the model to hallucinate new identities or backgrounds. By selectively refreshing the cache based on proximity, APR enables smooth, logically consistent transitions even when prompts share linguistic dependencies, significantly reducing visual jitter and maintaining narrative flow in interactive video synthesis.

%% file: sections/4_experiments.tex
\section{Experiments}
\subsection{Experiments Setup}
\subsubsection{\textbf{Implementation Details.}} 
We implement \textbf{Grounded Forcing} on the open-source Wan2.1-T2V-1.3B backbone~\cite{wan}, ensuring consistency with baseline methods including Self-Forcing~\cite{SelfForcing}, Rolling-Forcing~\cite{rollingforcing}, LongLive~\cite{longlive}, and Infinity-RoPE~\cite{InfinityRope}. The model natively generates 5-second videos at 16 FPS with a spatial resolution of $832 \times 480$. 

For the \textbf{Dual Memory KV Cache}, we configure the Global Consistency Memory (GCM) with a fixed size of 3 keyframes and the Local Temporal Memory (LTM) with a sliding window of 6 frames. This configuration balances long-term semantic anchoring with short-term motion fidelity. 
In the \textbf{Asymmetric Proximity Recache (APR)} module, the recache scaling factor $\alpha_t$ is linearly adjusted within the range $[0.0, 0.8]$ based on temporal proximity, ensuring smooth transitions without semantic shock. 
Our training protocol follows a two-stage strategy inspired by LongLive~\cite{longlive}. 
\textbf{Stage 1} focuses on standard generation stability, trained on 5-second clips for 1,200 steps using 32 NVIDIA H20 GPUs. 
\textbf{Stage 2} specializes in long-horizon consistency and interactive control, trained on sequences exceeding 5 seconds with dynamic prompt switching for 1,000 steps, also on 32 NVIDIA H20 GPUs. 
This progressive training ensures the model first masters basic coherence before adapting to complex interactive scenarios.

\subsubsection{\textbf{Baselines.}}  
We compare Grounded Forcing against representative autoregressive long-video generation models(all based on Wan2.1-T2V-1.3B for fairness), include LongLive~\cite{longlive}, Rolling Forcing \cite{rollingforcing}, $\infty$-Rope~\cite{InfinityRope}.

\subsubsection{\textbf{Evaluation.}} Following~\cite{rollingforcing, InfinityRope}, we employ VBench~\cite{huang2024vbench,zheng2025vbench} for quantitative evaluation. 
The assessment encompasses seven key metrics: subject consistency, background consistency, motion smoothness, temporal flickering, dynamic degree, aesthetic quality, and imaging quality. For the experiments, we randomly sample 100 prompts from MovieGenBench~\cite{moviegen} and generate videos across four durations: 5, 60, 120, and 240 seconds. For interactive evaluation, we used Qwen3-Max~\cite{yang2025qwen3} to generate 50 interactive prompt lists, each containing 6 sub-prompts. In our experiment, each sub-prompt generates a 10-second video clip, for a total of 60 seconds.

\subsection{Results}

\begin{table*}[t]
\centering
\caption{\textbf{Performance comparisons on 240s videos.} We evaluate seven metrics covering aesthetic quality, temporal consistency, and motion dynamics. Best and second-best results are \textbf{bolded} and \underline{underlined}, respectively.}
\vspace{-1em}
\label{tab:res_120s_240s}
\resizebox{\textwidth}{!}{
\begin{tabular}{l|ccccccc}
\toprule
\textbf{Model} &
\multicolumn{7}{c}{\textbf{Results on 240s $\uparrow$}} \\
\cmidrule(lr){2-8} 
&
\shortstack{\textbf{Aesthetic} \\ \textbf{Quality}} &
\shortstack{\textbf{Background} \\ \textbf{Consistency}} &
\shortstack{\textbf{Dynamic} \\ \textbf{Degree}} &
\shortstack{\textbf{Imaging} \\ \textbf{Quality}} &
\shortstack{\textbf{Motion} \\ \textbf{Smoothness}} &
\shortstack{\textbf{Subject} \\ \textbf{Consistency}} &
\shortstack{\textbf{Temporal} \\ \textbf{Flickering}} \\
\midrule
LongLive & 0.6032 & \underline{0.9208} & \textbf{0.60} & \underline{0.7245} & \textbf{0.9892} & 0.8973 & 0.9770 \\
Rolling Forcing & \textbf{0.6237} & 0.9144 & 0.53 & \textbf{0.7317} & \underline{0.9878} & 0.9034 & \textbf{0.9845} \\
Infinity Rope & 0.6048 & 0.9195 & \underline{0.58} & 0.7154 & 0.9873 & \underline{0.9066} & 0.9808 \\
\textbf{Ours} & \underline{0.6174} & \textbf{0.9265} & \textbf{0.60} & 0.7204 & \underline{0.9878} & \textbf{0.9163} & \underline{0.9812} \\
\bottomrule
\end{tabular}
}
\vspace{-1.3em}
\end{table*}

\subsubsection{Quantitative Results.}
\label{sec:qualitative}
We evaluate the performance of our method on generating 240-second videos, comparing against strong baselines including LongLive, Rolling Forcing, and Infinity-RoPE. As shown in Table \ref{tab:res_120s_240s}, we report results across seven metrics covering visual quality, temporal stability, and motion dynamics. Our method achieves the best performance in Background Consistency (0.9265) and Subject Consistency (0.9163), while ranking second in Aesthetic Quality and Temporal Flickering. Notably, our approach maintains a favorable balance between motion dynamics and stability, matching the highest Dynamic Degree (0.60) achieved by LongLive while significantly improving subject preservation. These results demonstrate that our method sustains high-quality generation over extended horizons, effectively mitigating the degradation and drift typically observed in long-video synthesis. For evaluations of the other three video lengths, please refer to the supplementary reference materials.

\begin{figure}[t]
    \centering
    \includegraphics[width=\linewidth]{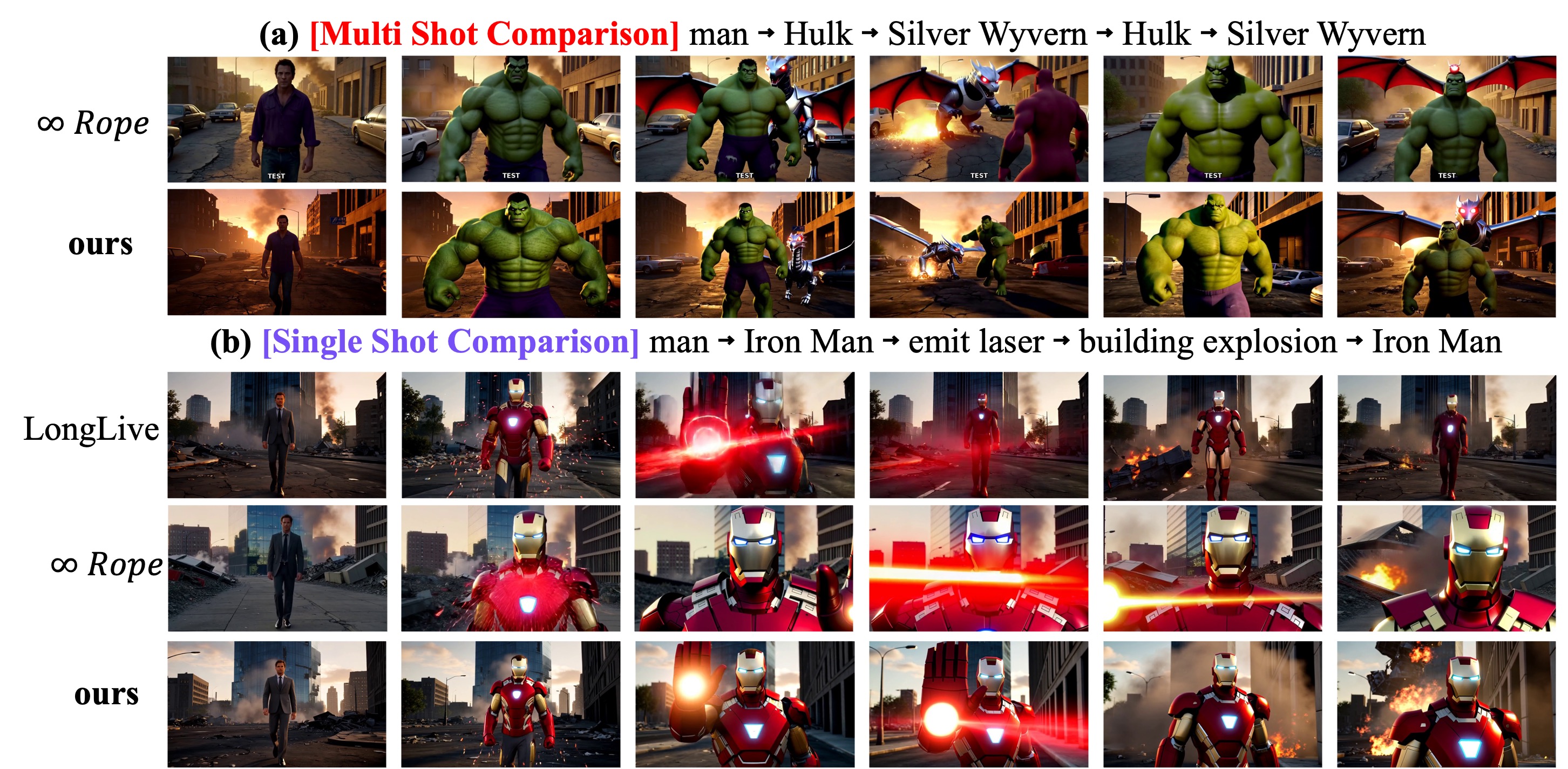}
    \caption{\textbf{Qualitative Comparison with Baselines.} 
    (a) \textbf{Multi-Shot Generation:} Our method maintains consistent identity through multiple semantic transitions (man $\rightarrow$ Hulk $\rightarrow$ Silver Wyvern), while Infinity-RoPE suffers from identity drift. 
    (b) \textbf{Single-Shot Generation:} Our approach preserves Iron Man's identity and action coherence (laser $\rightarrow$ explosion), whereas LongLive and Infinity-RoPE exhibit visual degradation.}
    \label{fig:comp1}
\end{figure}

\subsubsection{Qualitative Results.}
\textbf{Grounded Forcing}, as demonstrated in Figure~\ref{fig:comp1}, effectively overcomes the tendency of models to remain fixated on the first frame and enables the incorporation of new semantic content. In \textbf{Multi-Shot} generation (a), our method preserves identity across recurring states (man $\rightarrow$ Hulk $\rightarrow$ Silver Wyvern), while Infinity-RoPE suffers from severe forgetting. In Single-Shot sequence (b), we smoothly adapt to prompt switches (Iron Man $\rightarrow$ laser $\rightarrow$ explosion) without the abrupt visual jumps observed in baselines. Our Dual Memory, APR, and DR-RoPE work synergistically to maintain semantic consistency and visual fidelity throughout long-horizon generation.

\subsubsection{Ablation Studies}
To further validate the contribution of each component, we conduct an ablation study on a challenging multi-prompt interactive generation task. Specifically, we curate a set of 50 prompts with high transformation variations, generating 60-second sequences that involve 5 prompt switches to test the model's adaptability and memory retention. Table~\ref{tab:ablation} summarizes the results.
\begin{table*}[t]
\centering
\caption{\textbf{Ablation study on component effectiveness.} We evaluate the impact of removing key modules (Dual Mem, DR-RoPE, APR) on 60s multi-prompt generation. Best and second-best results are \textbf{bolded} and \underline{underlined}, respectively.}
\vspace{-1em}
\label{tab:ablation}
\resizebox{0.8\textwidth}{!}{
\begin{tabular}{l|cccc}
\toprule
\textbf{Model} &
\multicolumn{4}{c}{\textbf{Ablation Results on 60s $\uparrow$}} \\
\cmidrule(lr){2-5} 
&
\shortstack{\textbf{Aesthetic} \\ \textbf{Quality}} &
\shortstack{\textbf{Background} \\ \textbf{Consistency}} &
\shortstack{\textbf{Subject} \\ \textbf{Consistency}} &
\shortstack{\textbf{Temporal} \\ \textbf{Flickering}} \\
\midrule
w/o Dual Mem & \textbf{0.6473} & 0.8467 & 0.7322 & \textbf{0.9734} \\
w/o DR-RoPE  & 0.6278 & \underline{0.8729} & 0.7606 & 0.9716 \\
w/o APR      & 0.6386 & 0.8603 & \underline{0.7719} & 0.9636 \\
\textbf{Ours (full)} & \underline{0.6440} & \textbf{0.8795} & \textbf{0.7770} & \underline{0.9723} \\
\bottomrule
\end{tabular}
}
\vspace{-1.3em}
\end{table*}

\subsection{User Study}

We conducted a user study with 10 participants using a 5-point Likert scale (1–5) to evaluate \textbf{Subject Consistency}, \textbf{Background Consistency}, \textbf{Aesthetic Quality}, and \textbf{Text Adherence}. Table \ref{tab:user_study_results} summarizes the mean scores. Our method demonstrates superior performance, securing the top rank in all four categories. Specifically, we observe a significant margin over the current state-of-the-art methods in Subject Consistency and Background Consistency. These findings confirm that our approach not only improves quantitative metrics but also enhances the subjective viewing experience regarding temporal stability and content fidelity.

\vspace{-5pt}
\begin{table*}[ht]
    \centering
    \caption{\textbf{User study results on 5-point Likert scale.} We compare our method with LongLive and Infinity-Rope across four perceptual metrics. Higher scores indicate better performance. Best results are in \textbf{bold}.}
    \label{tab:user_study_results}
    \begin{tabular}{lcccc}
        \toprule
        Metric & LongLive & Infinity-Rope & Ours \\
        \midrule
        Subject Consistency   & 3.02 & 2.98 & \textbf{3.66} \\
        Background Consistency & 3.04 & 3.08 & \textbf{3.78} \\
        Aesthetic Quality     & 3.45 & 3.34 & \textbf{3.42} \\
        Text Adherence        & 3.12 & 3.15 & \textbf{3.72} \\
        \bottomrule
    \end{tabular}
    \vspace{-15pt}
\end{table*}

\vspace{-10pt}

%% file: sections/5_conclusion.tex
\section{Conclusion}

In this work, we present \textbf{Grounded Forcing}, a comprehensive framework for interactive long-horizon video generation that systematically addresses the entangled challenges of semantic forgetting, visual drift, and controllability degradation. By decoupling time-independent semantics from proximal dynamics through our Dual Memory KV Cache, Dual-Reference RoPE Injection, and Asymmetric Proximity Recache, we enable stable autoregressive synthesis over extended temporal horizons. Our approach anchors global identity features while accommodating flexible local motion and smooth prompt transitions, transforming standard causal models into robust engines for immersive simulation.

Extensive experiments demonstrate that Grounded Forcing achieves state-of-the-art semantic consistency and visual stability in long-form video generation while maintaining real-time performance. We hope this work inspires further research into memory-efficient architectures for infinite-horizon generative models, paving the way toward truly interactive world simulators.

%% file: sections/X_supp.tex
\clearpage
\appendix
\begin{center}
    {\Large \bfseries Supplementary Material}
\end{center}

\setcounter{section}{0}
\renewcommand{\thesection}{\Alph{section}}

\section{Additional Quantitative Results}

We provide comprehensive evaluations across three video durations (5s, 60s, and 240s) to demonstrate the scalability and robustness of \textbf{Grounded Forcing}. While the main paper reports 240s results, we include the remaining settings below: on 5s videos (Table~\ref{tab:res_5s}), our approach maintains competitive or superior results—particularly in Subject Consistency (0.9421) and Background Consistency (0.9612)—validating Dual Memory effectiveness even for brief sequences; on 60s videos (Table~\ref{tab:res_60s}), baseline methods exhibit noticeable degradation in semantic consistency as temporal horizons extend, while our method preserves clear advantages in Subject Consistency (0.9317) and Background Consistency (0.9356), demonstrating Global Consistency Memory's capacity to maintain semantic anchors over extended durations.

\begin{table*}[h]
\centering
\vspace{-10pt}
\caption{\textbf{Performance comparisons on 5s videos.} All methods perform well at short horizons; our approach maintains advantages in semantic consistency metrics. Best and second-best results are \textbf{bolded} and \underline{underlined}, respectively.}
\vspace{-10pt}
\label{tab:res_5s}
\resizebox{\textwidth}{!}{
\begin{tabular}{l|ccccccc}
\toprule
\textbf{Model} &
\multicolumn{7}{c}{\textbf{Results on 5s $\uparrow$}} \\
\cmidrule(lr){2-8} 
&
\shortstack{\textbf{Aesthetic} \\ \textbf{Quality}} &
\shortstack{\textbf{Background} \\ \textbf{Consistency}} &
\shortstack{\textbf{Dynamic} \\ \textbf{Degree}} &
\shortstack{\textbf{Imaging} \\ \textbf{Quality}} &
\shortstack{\textbf{Motion} \\ \textbf{Smoothness}} &
\shortstack{\textbf{Subject} \\ \textbf{Consistency}} &
\shortstack{\textbf{Temporal} \\ \textbf{Flickering}} \\
\midrule
LongLive & 0.6512 & \underline{0.9534} & \textbf{0.62} & \underline{0.7823} & \textbf{0.9934} & 0.9412 & \underline{0.9923} \\
Rolling Forcing & \textbf{0.6689} & 0.9487 & 0.58 & \textbf{0.7891} & \underline{0.9928} & 0.9387 & \textbf{0.9931} \\
Infinity Rope & 0.6445 & 0.9501 & \underline{0.60} & 0.7756 & 0.9921 & \textbf{0.9445} & 0.9918 \\
\textbf{Ours} & \underline{0.6623} & \textbf{0.9612} & \underline{0.60} & 0.7801 & 0.9924 & \underline{0.9421} & \textbf{0.9931} \\
\bottomrule
\end{tabular}
}
\vspace{-30pt}
\end{table*}

\begin{table*}[h]
\centering
\caption{\textbf{Performance comparisons on 60s videos.} Our method shows growing advantages in consistency metrics as generation horizon extends. Best and second-best results are \textbf{bolded} and \underline{underlined}, respectively.}
\vspace{-10pt}
\label{tab:res_60s}
\resizebox{\textwidth}{!}{
\begin{tabular}{l|ccccccc}
\toprule
\textbf{Model} &
\multicolumn{7}{c}{\textbf{Results on 60s $\uparrow$}} \\
\cmidrule(lr){2-8} 
&
\shortstack{\textbf{Aesthetic} \\ \textbf{Quality}} &
\shortstack{\textbf{Background} \\ \textbf{Consistency}} &
\shortstack{\textbf{Dynamic} \\ \textbf{Degree}} &
\shortstack{\textbf{Imaging} \\ \textbf{Quality}} &
\shortstack{\textbf{Motion} \\ \textbf{Smoothness}} &
\shortstack{\textbf{Subject} \\ \textbf{Consistency}} &
\shortstack{\textbf{Temporal} \\ \textbf{Flickering}} \\
\midrule
LongLive & 0.6145 & 0.9289 & \textbf{0.61} & \underline{0.7534} & \textbf{0.9912} & 0.9123 & 0.9834 \\
Rolling Forcing & \textbf{0.6378} & 0.9267 & 0.56 & \textbf{0.7612} & \underline{0.9901} & \underline{0.9256} & \textbf{0.9867} \\
Infinity Rope & 0.6189 & \underline{0.9313} & \underline{0.59} & 0.7445 & 0.9895 & 0.9234 & \underline{0.9845} \\
\textbf{Ours} & \underline{0.6298} & \textbf{0.9356} & \textbf{0.61} & 0.7489 & 0.9900 & \textbf{0.9317} & 0.9841 \\
\bottomrule
\end{tabular}
}

\end{table*}

\section{Additional Qualitative Results}
\label{sec:additional_qualitative}

We provide some comprehensive visualizations demonstrating the effectiveness of \mbox{Grounded Forcing} in maintaining long-term consistency across various generation scenarios. All videos are generated using the Wan2.1-T2V-1.3B backbone integrated with our proposed framework. The following examples highlight our model's capability to suppress visual drift, preserve identity, and handle dynamic semantic transitions.

\begin{figure}[ht!]
    \centering
    \includegraphics[width=\linewidth]{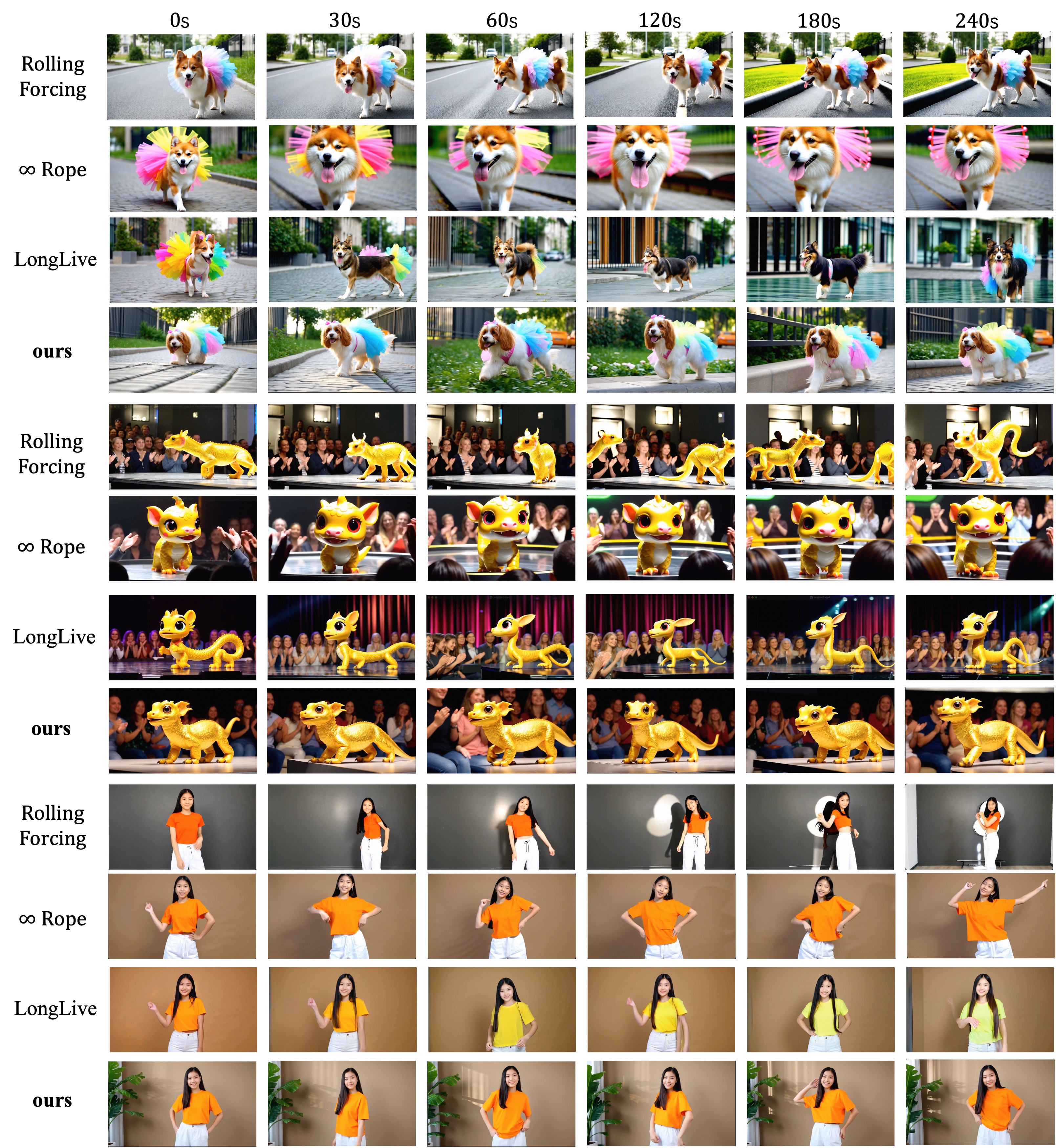}
    \caption{\textbf{Single-prompt generation (240s).} Our method demonstrates stronger subject-matter and background consistency in long-range generation results compared to other methods.}
    \label{fig:single_prompt}
\end{figure}

\paragraph{Single-Prompt Long-Horizon Generation (240s).}
Figure~\ref{fig:single_prompt} demonstrates stable single-prompt generation up to \textbf{240 seconds}. By preserving global semantic anchors via Dual Memory KV Cache and preventing positional drift with Dual-Reference RoPE Injection, our method maintains subject identity and background coherence throughout.

\begin{figure}[ht!]
    \centering
    \vspace{-10pt}
    \includegraphics[width=\linewidth]{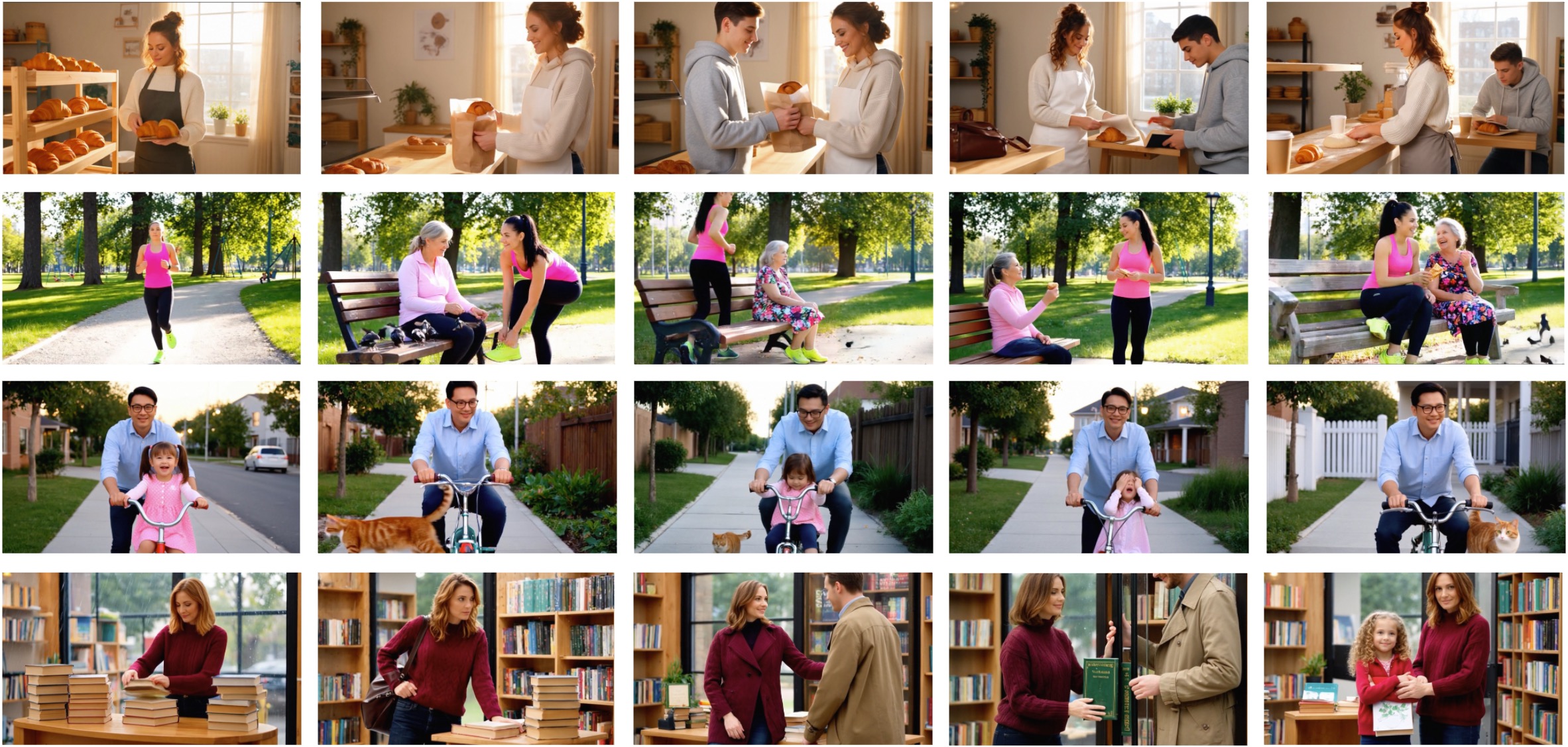}
    \vspace{-10pt}
    \caption{\textbf{Multi-shot generation with narrative continuity (60s).} Character identity (highlighted in red boxes) remains consistent across diverse scenes and camera angles. The GCM successfully anchors core semantic features while allowing local dynamics to adapt to new prompts via APR and LTM Reset.}
    \label{fig:multi_shot}
\end{figure}

\begin{figure}[ht!]
    \centering
    \includegraphics[width=\linewidth]{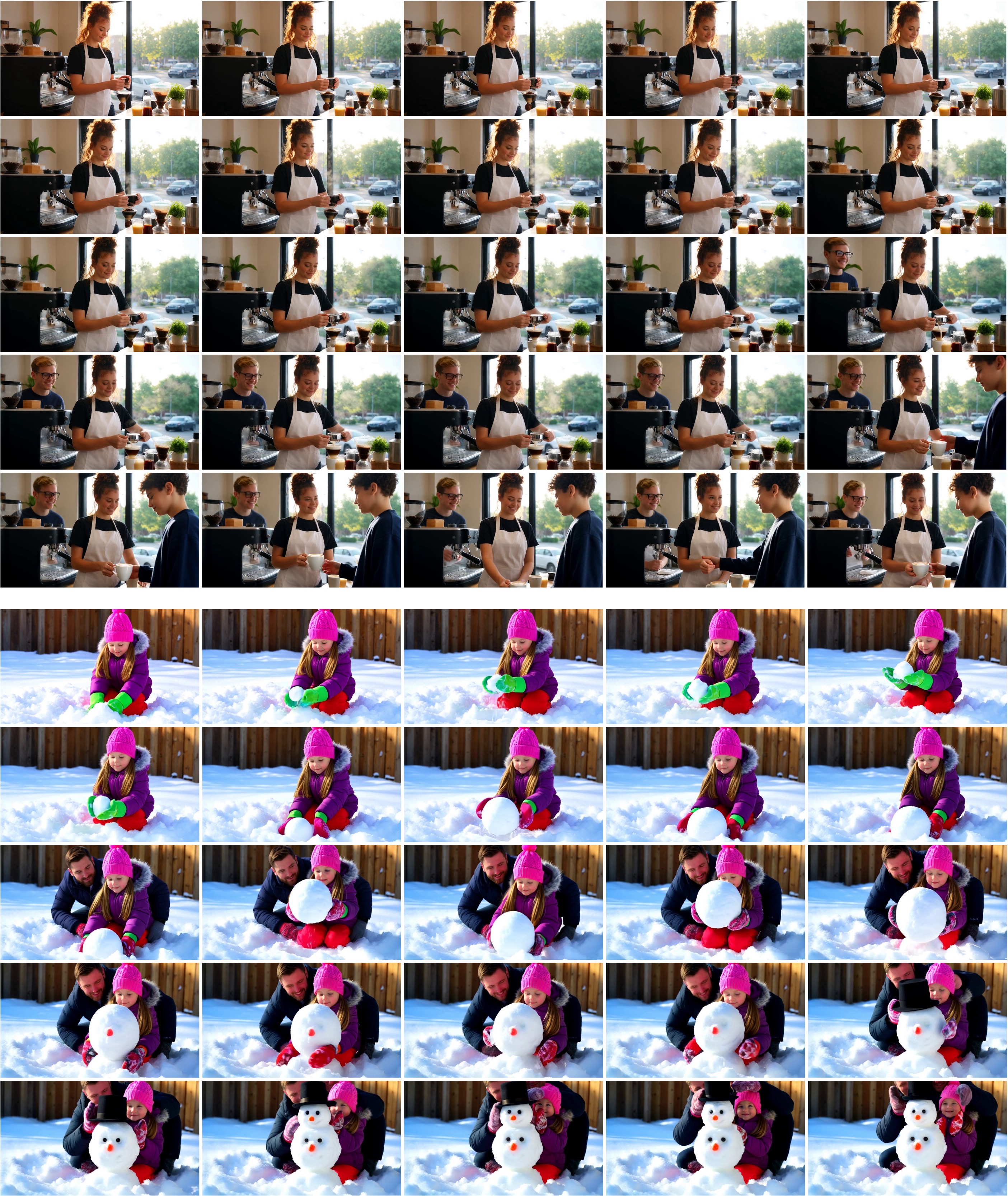}
    \caption{\textbf{Interactive Prompt Switching (60s).} Our method demonstrates stronger consistency of main content and smoother video transitions in long-range interactive prompt generation results.}
    \label{fig:single_shot_1}
\end{figure}

\begin{figure}[ht!]
    \centering
    \includegraphics[width=\linewidth]{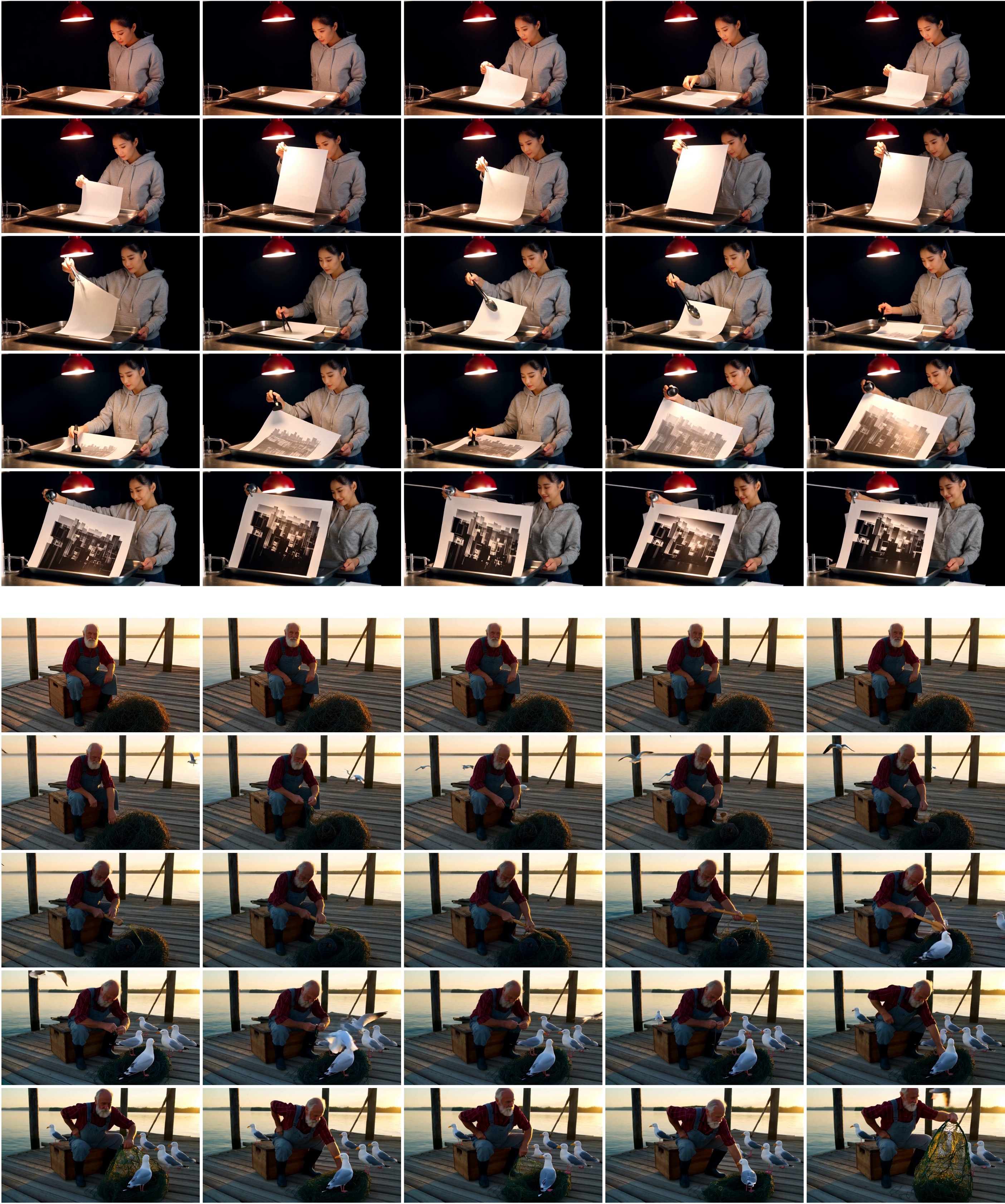}
    \caption{\textbf{Interactive Prompt Switching (60s).} Our method demonstrates stronger consistency of main content and smoother video transitions in long-range interactive prompt generation results.}
    \label{fig:single_shot_2}
\end{figure}

\paragraph{Interactive Prompt Switching and Multi-Shot Generation (60s).}
Figures~\ref{fig:single_shot_1} and \ref{fig:single_shot_2} demonstrate robust multi-prompt generation over a \textbf{60-second} horizon (6 sub-prompts). GCMnanchors core identity across transitions, while APR ensures smooth semantic inheritance. Our method successfully adapts to interactive instructions while maintaining coherence. Additionally, Figure~\ref{fig:multi_shot} illustrates cinematic multi-shot generation, where local memory reset  enables clean scene cuts without temporal contamination, preserving character identity across diverse scenes.

%% file: main.bib
@String(AAAI  = {AAAI})

@inproceedings{Dit,
  title={Scalable diffusion models with transformers},
  author={Peebles, William and Xie, Saining},
  booktitle={Proceedings of the IEEE/CVF international conference on computer vision},
  pages={4195--4205},
  year={2023}
}

@article{wan,
  title={Wan: Open and advanced large-scale video generative models},
  author={Wan, Team and Wang, Ang and Ai, Baole and Wen, Bin and Mao, Chaojie and Xie, Chen-Wei and Chen, Di and Yu, Feiwu and Zhao, Haiming and Yang, Jianxiao and others},
  journal={arXiv preprint arXiv:2503.20314},
  year={2025}
}

@inproceedings{CausVid,
  title={From slow bidirectional to fast autoregressive video diffusion models},
  author={Yin, Tianwei and Zhang, Qiang and Zhang, Richard and Freeman, William T and Durand, Fredo and Shechtman, Eli and Huang, Xun},
  booktitle={Proceedings of the IEEE/CVF Conference on Computer Vision and Pattern Recognition},
  pages={22963--22974},
  year={2025}
}

@article{SelfForcing,
  title={Self forcing: Bridging the train-test gap in autoregressive video diffusion},
  author={Huang, Xun and Li, Zhengqi and He, Guande and Zhou, Mingyuan and Shechtman, Eli},
  journal={arXiv preprint arXiv:2506.08009},
  year={2025}
}

@article{InfinityRope,
  title={Infinity-rope: Action-controllable infinite video generation emerges from autoregressive self-rollout},
  author={Yesiltepe, Hidir and Meral, Tuna Han Salih and Akan, Adil Kaan and Oktay, Kaan and Yanardag, Pinar},
  journal={arXiv preprint arXiv:2511.20649},
  year={2025}
}

@article{framepack,
  title={Packing input frame context in next-frame prediction models for video generation},
  author={Zhang, Lvmin and Agrawala, Maneesh},
  journal={arXiv e-prints},
  pages={arXiv--2504},
  year={2025}
}

@article{rollingforcing,
  title={Rolling forcing: Autoregressive long video diffusion in real time},
  author={Liu, Kunhao and Hu, Wenbo and Xu, Jiale and Shan, Ying and Lu, Shijian},
  journal={arXiv preprint arXiv:2509.25161},
  year={2025}
}

@article{longlive,
  title={Longlive: Real-time interactive long video generation},
  author={Yang, Shuai and Huang, Wei and Chu, Ruihang and Xiao, Yicheng and Zhao, Yuyang and Wang, Xianbang and Li, Muyang and Xie, Enze and Chen, Yingcong and Lu, Yao and others},
  journal={arXiv preprint arXiv:2509.22622},
  year={2025}
}

@article{hunyuanvideo,
  title={Hunyuanvideo: A systematic framework for large video generative models},
  author={Kong, Weijie and Tian, Qi and Zhang, Zijian and Min, Rox and Dai, Zuozhuo and Zhou, Jin and Xiong, Jiangfeng and Li, Xin and Wu, Bo and Zhang, Jianwei and others},
  journal={arXiv preprint arXiv:2412.03603},
  year={2024}
}

@article{freelong,
  title={Freelong: Training-free long video generation with spectralblend temporal attention},
  author={Lu, Yu and Liang, Yuanzhi and Zhu, Linchao and Yang, Yi},
  journal={Advances in Neural Information Processing Systems},
  volume={37},
  pages={131434--131455},
  year={2024}
}

@article{riflex,
  title={Riflex: A free lunch for length extrapolation in video diffusion transformers},
  author={Zhao, Min and He, Guande and Chen, Yixiao and Zhu, Hongzhou and Li, Chongxuan and Zhu, Jun},
  journal={arXiv preprint arXiv:2502.15894},
  year={2025}
}

@article{SelfForcing++,
  title={Self-forcing++: Towards minute-scale high-quality video generation},
  author={Cui, Justin and Wu, Jie and Li, Ming and Yang, Tao and Li, Xiaojie and Wang, Rui and Bai, Andrew and Ban, Yuanhao and Hsieh, Cho-Jui},
  journal={arXiv preprint arXiv:2510.02283},
  year={2025}
}

@article{moviegen,
  title={Movie gen: A cast of media foundation models},
  author={Polyak, Adam and Zohar, Amit and Brown, Andrew and Tjandra, Andros and Sinha, Animesh and Lee, Ann and Vyas, Apoorv and Shi, Bowen and Ma, Chih-Yao and Chuang, Ching-Yao and others},
  journal={arXiv preprint arXiv:2410.13720},
  year={2024}
}

@article{seawead,
  title={Seaweed-7b: Cost-effective training of video generation foundation model},
  author={Seawead, Team and Yang, Ceyuan and Lin, Zhijie and Zhao, Yang and Lin, Shanchuan and Ma, Zhibei and Guo, Haoyuan and Chen, Hao and Qi, Lu and Wang, Sen and others},
  journal={arXiv preprint arXiv:2504.08685},
  year={2025}
}

@article{cogvideox,
  title={Cogvideox: Text-to-video diffusion models with an expert transformer},
  author={Yang, Zhuoyi and Teng, Jiayan and Zheng, Wendi and Ding, Ming and Huang, Shiyu and Xu, Jiazheng and Yang, Yuanming and Hong, Wenyi and Zhang, Xiaohan and Feng, Guanyu and others},
  journal={arXiv preprint arXiv:2408.06072},
  year={2024}
}

@article{ltx,
  title={Ltx-video: Realtime video latent diffusion},
  author={HaCohen, Yoav and Chiprut, Nisan and Brazowski, Benny and Shalem, Daniel and Moshe, Dudu and Richardson, Eitan and Levin, Eran and Shiran, Guy and Zabari, Nir and Gordon, Ori and others},
  journal={arXiv preprint arXiv:2501.00103},
  year={2024}
}

@inproceedings{huang2024vbench,
  title={Vbench: Comprehensive benchmark suite for video generative models},
  author={Huang, Ziqi and He, Yinan and Yu, Jiashuo and Zhang, Fan and Si, Chenyang and Jiang, Yuming and Zhang, Yuanhan and Wu, Tianxing and Jin, Qingyang and Chanpaisit, Nattapol and others},
  booktitle={Proceedings of the IEEE/CVF Conference on Computer Vision and Pattern Recognition},
  pages={21807--21818},
  year={2024}
}

@article{contextforcing,
  title={Context Forcing: Consistent Autoregressive Video Generation with Long Context},
  author={Chen, Shuo and Wei, Cong and Sun, Sun and Nie, Ping and Zhou, Kai and Zhang, Ge and Yang, Ming-Hsuan and Chen, Wenhu},
  journal={arXiv preprint arXiv:2602.06028},
  year={2026}
}

@article{su2024roformer,
  title={Roformer: Enhanced transformer with rotary position embedding},
  author={Su, Jianlin and Ahmed, Murtadha and Lu, Yu and Pan, Shengfeng and Bo, Wen and Liu, Yunfeng},
  journal={Neurocomputing},
  volume={568},
  pages={127063},
  year={2024},
  publisher={Elsevier}
}

@article{teng2025magi,
  title={Magi-1: Autoregressive video generation at scale},
  author={Teng, Hansi and Jia, Hongyu and Sun, Lei and Li, Lingzhi and Li, Maolin and Tang, Mingqiu and Han, Shuai and Zhang, Tianning and Zhang, WQ and Luo, Weifeng and others},
  journal={arXiv preprint arXiv:2505.13211},
  year={2025}
}

@article{chen2025skyreels,
  title={Skyreels-v2: Infinite-length film generative model},
  author={Chen, Guibin and Lin, Dixuan and Yang, Jiangping and Lin, Chunze and Zhu, Junchen and Fan, Mingyuan and Zhang, Hao and Chen, Sheng and Chen, Zheng and Ma, Chengcheng and others},
  journal={arXiv preprint arXiv:2504.13074},
  year={2025}
}

@article{ho2020denoising,
  title={Denoising diffusion probabilistic models},
  author={Ho, Jonathan and Jain, Ajay and Abbeel, Pieter},
  journal={Advances in neural information processing systems},
  volume={33},
  pages={6840--6851},
  year={2020}
}

@article{lipman2022flow,
  title={Flow matching for generative modeling},
  author={Lipman, Yaron and Chen, Ricky TQ and Ben-Hamu, Heli and Nickel, Maximilian and Le, Matt},
  journal={arXiv preprint arXiv:2210.02747},
  year={2022}
}

@inproceedings{dmd,
  title={One-step diffusion with distribution matching distillation},
  author={Yin, Tianwei and Gharbi, Micha{\"e}l and Zhang, Richard and Shechtman, Eli and Durand, Fredo and Freeman, William T and Park, Taesung},
  booktitle={Proceedings of the IEEE/CVF conference on computer vision and pattern recognition},
  pages={6613--6623},
  year={2024}
}

@article{dmd2,
  title={Improved distribution matching distillation for fast image synthesis},
  author={Yin, Tianwei and Gharbi, Micha{\"e}l and Park, Taesung and Zhang, Richard and Shechtman, Eli and Durand, Fredo and Freeman, Bill},
  journal={Advances in neural information processing systems},
  volume={37},
  pages={47455--47487},
  year={2024}
}

@article{diffusionforcing,
  title={Diffusion forcing: Next-token prediction meets full-sequence diffusion},
  author={Chen, Boyuan and Mart{\'\i} Mons{\'o}, Diego and Du, Yilun and Simchowitz, Max and Tedrake, Russ and Sitzmann, Vincent},
  journal={Advances in Neural Information Processing Systems},
  volume={37},
  pages={24081--24125},
  year={2024}
}

@article{zhang2025pretraining,
  title={Pretraining Frame Preservation in Autoregressive Video Memory Compression},
  author={Zhang, Lvmin and Cai, Shengqu and Li, Muyang and Zeng, Chong and Lu, Beijia and Rao, Anyi and Han, Song and Wetzstein, Gordon and Agrawala, Maneesh},
  journal={arXiv preprint arXiv:2512.23851},
  year={2025}
}

@article{lu2025reward,
  title={Reward forcing: Efficient streaming video generation with rewarded distribution matching distillation},
  author={Lu, Yunhong and Zeng, Yanhong and Li, Haobo and Ouyang, Hao and Wang, Qiuyu and Cheng, Ka Leong and Zhu, Jiapeng and Cao, Hengyuan and Zhang, Zhipeng and Zhu, Xing and others},
  journal={arXiv preprint arXiv:2512.04678},
  year={2025}
}

@article{guo2025end,
  title={End-to-end training for autoregressive video diffusion via self-resampling},
  author={Guo, Yuwei and Yang, Ceyuan and He, Hao and Zhao, Yang and Wei, Meng and Yang, Zhenheng and Huang, Weilin and Lin, Dahua},
  journal={arXiv preprint arXiv:2512.15702},
  year={2025}
}

@article{yang2025qwen3,
  title={Qwen3 technical report},
  author={Yang, An and Li, Anfeng and Yang, Baosong and Zhang, Beichen and Hui, Binyuan and Zheng, Bo and Yu, Bowen and Gao, Chang and Huang, Chengen and Lv, Chenxu and others},
  journal={arXiv preprint arXiv:2505.09388},
  year={2025}
}

@article{zheng2025vbench,
  title={Vbench-2.0: Advancing video generation benchmark suite for intrinsic faithfulness},
  author={Zheng, Dian and Huang, Ziqi and Liu, Hongbo and Zou, Kai and He, Yinan and Zhang, Fan and Gu, Lulu and Zhang, Yuanhan and He, Jingwen and Zheng, Wei-Shi and others},
  journal={arXiv preprint arXiv:2503.21755},
  year={2025}
}

@article{chen2026conceptweaver,
  title={ConceptWeaver: Weaving Disentangled Concepts with Flow},
  author={Chen, Jintao and Hao, Aiming and Chen, Xiaoqing and Bai, Chengyu and Chen, Chubin and Li, Yanxun and Wu, Jiahong and Chu, Xiangxiang and Zhang, Shanghang},
  journal={arXiv preprint arXiv:2603.28493},
  year={2026}
}

@article{bai2025uniedit,
  title={UniEdit-I: Training-free Image Editing for Unified VLM via Iterative Understanding, Editing and Verifying},
  author={Bai, Chengyu and Chen, Jintao and Bai, Xiang and Chen, Yilong and She, Qi and Lu, Ming and Zhang, Shanghang},
  journal={arXiv preprint arXiv:2508.03142},
  year={2025}
}

@inproceedings{mao2026omni,
  title={Omni-effects: Unified and spatially-controllable visual effects generation},
  author={Mao, Fangyuan and Hao, Aiming and Chen, Jintao and Liu, Dongxia and Feng, Xiaokun and Zhu, Jiashu and Wu, Meiqi and Chen, Chubin and Wu, Jiahong and Chu, Xiangxiang},
  booktitle={Proceedings of the AAAI Conference on Artificial Intelligence},
  volume={40},
  number={10},
  pages={7927--7935},
  year={2026}
}
